\documentclass[sigconf]{acmart}

\AtBeginDocument{%
  \providecommand\BibTeX{{%
    \normalfont B\kern-0.5em{\scshape i\kern-0.25em b}\kern-0.8em\TeX}}}

\copyrightyear{2021} 
\acmYear{2021} 
\setcopyright{acmcopyright}\acmConference[MM '21]{Proceedings of the 29th ACM International Conference on Multimedia}{October 20--24, 2021}{Virtual Event, China}
\acmBooktitle{Proceedings of the 29th ACM International Conference on Multimedia (MM '21), October 20--24, 2021, Virtual Event, China}
\acmPrice{15.00}
\acmDOI{10.1145/3474085.3475345}
\acmISBN{978-1-4503-8651-7/21/10}

\usepackage{multirow}
\usepackage{graphicx}
\usepackage{balance}
\usepackage{caption}
\usepackage{subcaption}

\begin{document}

\title{StrucTexT: Structured Text Understanding with Multi-Modal Transformers}

\author{Yulin Li}
\authornote{Equal contribution. This work is done when Yuxi Qian is an intern at Baidu Inc.}

\affiliation{
  \institution{Department of Computer Vision \\ Technology (VIS), Baidu Inc.}
  \streetaddress{}
  \city{}
  \country{}}
\email{liyulin03@baidu.com}
\author{Yuxi Qian}
\authornotemark[1]
\affiliation{
  \institution{Beijing University of Posts \\ and Telecommunications}
  \streetaddress{}
  \city{}
  \country{}}
\email{qianyuxi@bupt.edu.cn}

\author{Yuechen Yu}
\authornotemark[1]
\affiliation{
  \institution{Department of Computer Vision \\ Technology (VIS), Baidu Inc.}
  \streetaddress{}
  \city{}
  \country{}}
\email{yuyuechen@baidu.com}

\author{Xiameng Qin}
\affiliation{
  \institution{Department of Computer Vision \\ Technology (VIS), Baidu Inc.}
  \streetaddress{}
  \city{}
  \country{}}
\email{qinxiameng@baidu.com}

\author{Chengquan Zhang}
\authornote{Corresponding author.}
\affiliation{
  \institution{Department of Computer Vision \\ Technology (VIS), Baidu Inc.}
  \streetaddress{}
  \city{}
  \country{}}
\email{zhangchengquan@baidu.com}

\author{Yan Liu}
\affiliation{
  \institution{Taikang Insurance Group}
  \streetaddress{}
  \city{}
  \country{}}
\email{liuyan146@taikanglife.com}

\author{Kun Yao, Junyu Han}
% \author{Junyu Han}
% \email{hanjunyu@baidu.com}
\affiliation{
  \institution{Department of Computer Vision \\ Technology (VIS), Baidu Inc.}
  \streetaddress{}
  \city{}
  \country{}}
\email{{yaokun01, hanjunyu}@baidu.com}

\author{Jingtuo Liu, Errui Ding}
% \author{Errui Ding}
% \email{dingerrui@baidu.com}
\affiliation{
  \institution{Department of Computer Vision \\ Technology (VIS), Baidu Inc.}
  \streetaddress{}
  \city{}
  \country{}}
\email{{liujingtuo, dingerrui}@baidu.com}

\begin{abstract}
Structured text understanding on Visually Rich Documents (VRDs) is a crucial part of Document Intelligence. Due to the complexity of content and layout in VRDs, structured text understanding has been a challenging task. Most existing studies decoupled this problem into two sub-tasks: \textit{entity labeling} and \textit{entity linking}, which require an entire understanding of the context of documents at both token and segment levels. However, little work has been concerned with the solutions that efficiently extract the structured data from different levels. This paper proposes a unified framework named \textbf{StrucTexT}, which is flexible and effective for handling both sub-tasks. Specifically, based on the transformer, we introduce a segment-token aligned encoder to deal with the entity labeling and entity linking tasks at different levels of granularity. Moreover, we design a novel pre-training strategy with three self-supervised tasks to learn a richer representation. StrucTexT uses the existing Masked Visual Language Modeling task and the new Sentence Length Prediction and Paired Boxes Direction tasks to incorporate the multi-modal information across text, image, and layout. We evaluate our method for structured text understanding at segment-level and token-level and show it outperforms the state-of-the-art counterparts with significantly superior performance on the FUNSD, SROIE, and EPHOIE datasets.

\end{abstract}




\maketitle

\renewcommand{\shortauthors}{Li and Qian, et al.}
\fancyhead{}

\section{Introduction}
Understanding the structured document is a critical component of document intelligence that automatically explores the structured text information from the Visually Rich Documents (VRDs) such as forms, receipts, invoices, etc. Such task aims to extract the key information of text fields and the links among the semantic entities from VRDs, which named entity labeling and entity linking tasks~\cite{jaume2019funsd} respectively. Structured text understanding has attracted increasing attention in both academia and industry. In reality, it plays a crucial role in developing digital transformation processes in office automation, accounting systems, and electronically archived. It offers businesses significant time savings on processing the million of forms and invoices every day.

\begin{figure*}[t]
	\centering
	\subfloat[Token-based Entity Labeling]{\includegraphics[height=0.41\linewidth, trim=0 0 16.7cm 0, clip]{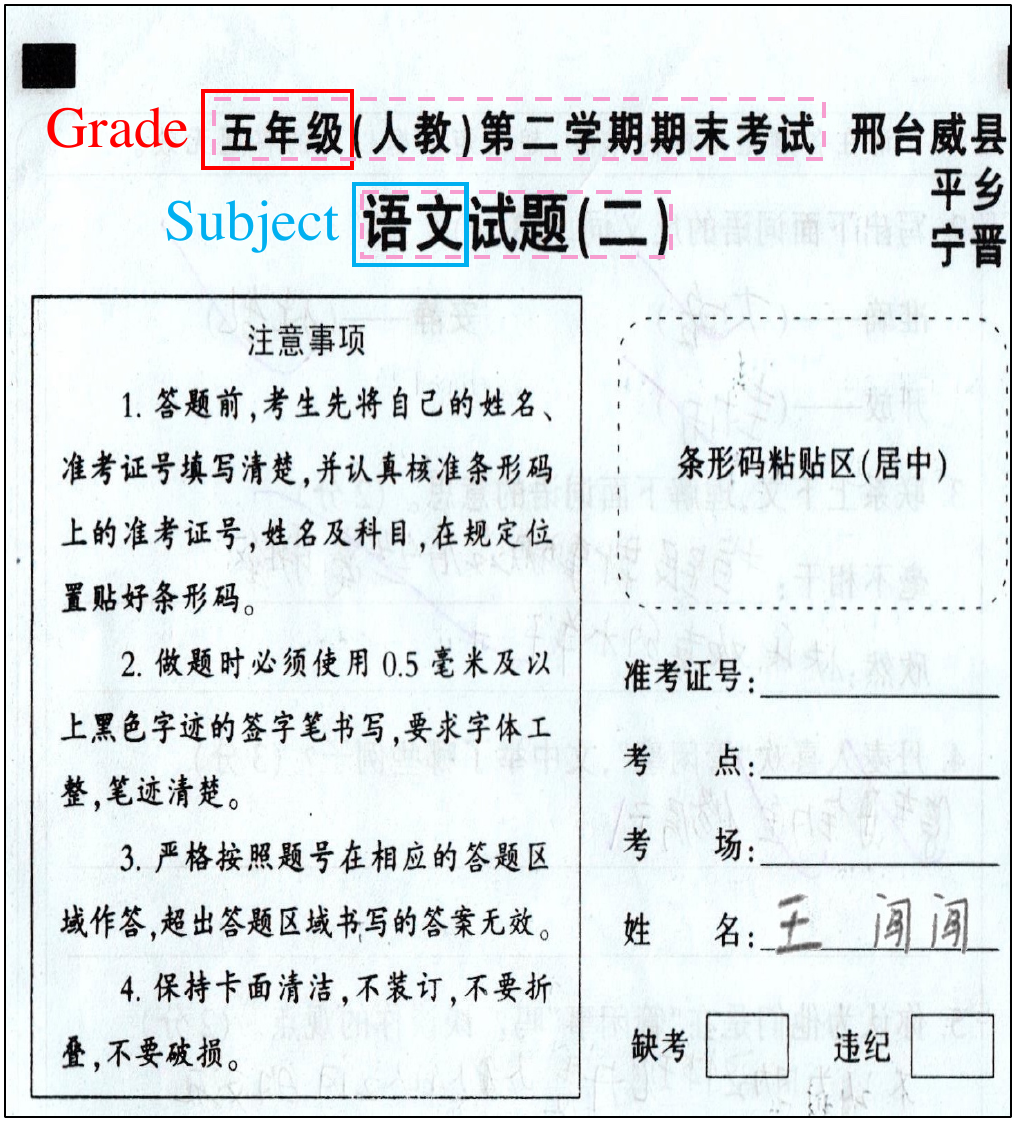}\label{fig:token_labeling}}
	\subfloat[Segment-based Entity Labeling]{\includegraphics[height=0.41\linewidth, trim=0 0 21.1cm 0, clip]{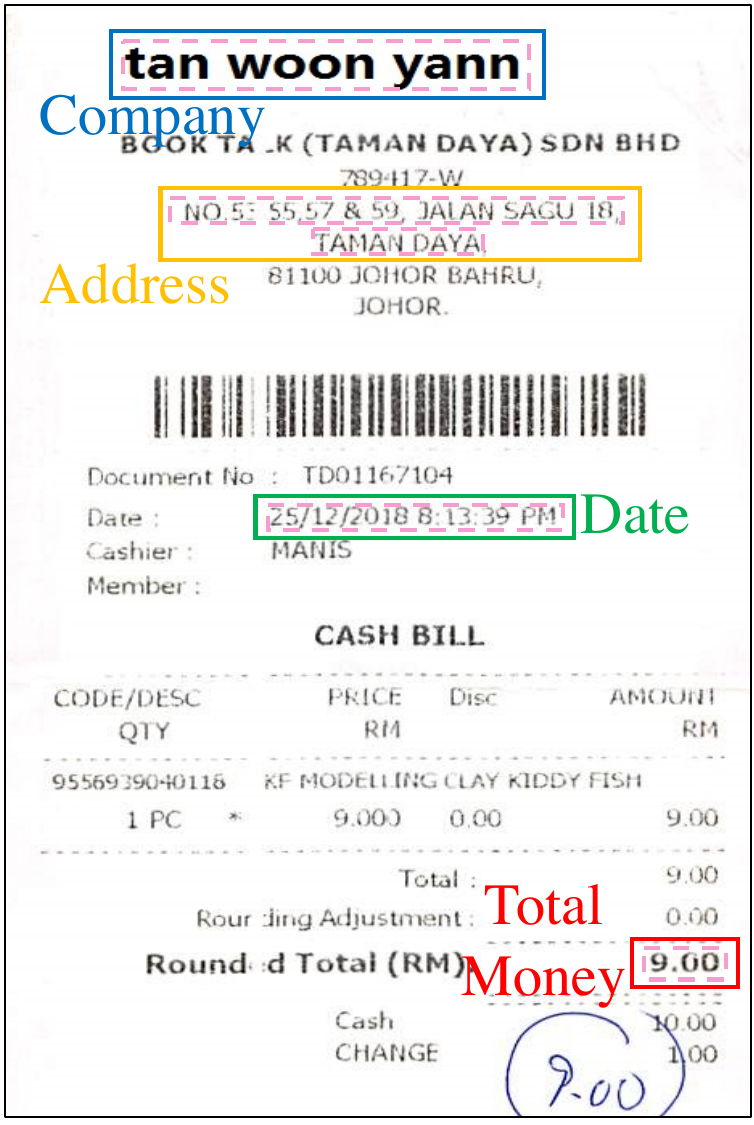}\label{fig:segment_labeling}}
	\subfloat[Entity Linking]{\includegraphics[height=0.41\linewidth, trim=0 0 17.7cm 0, clip]{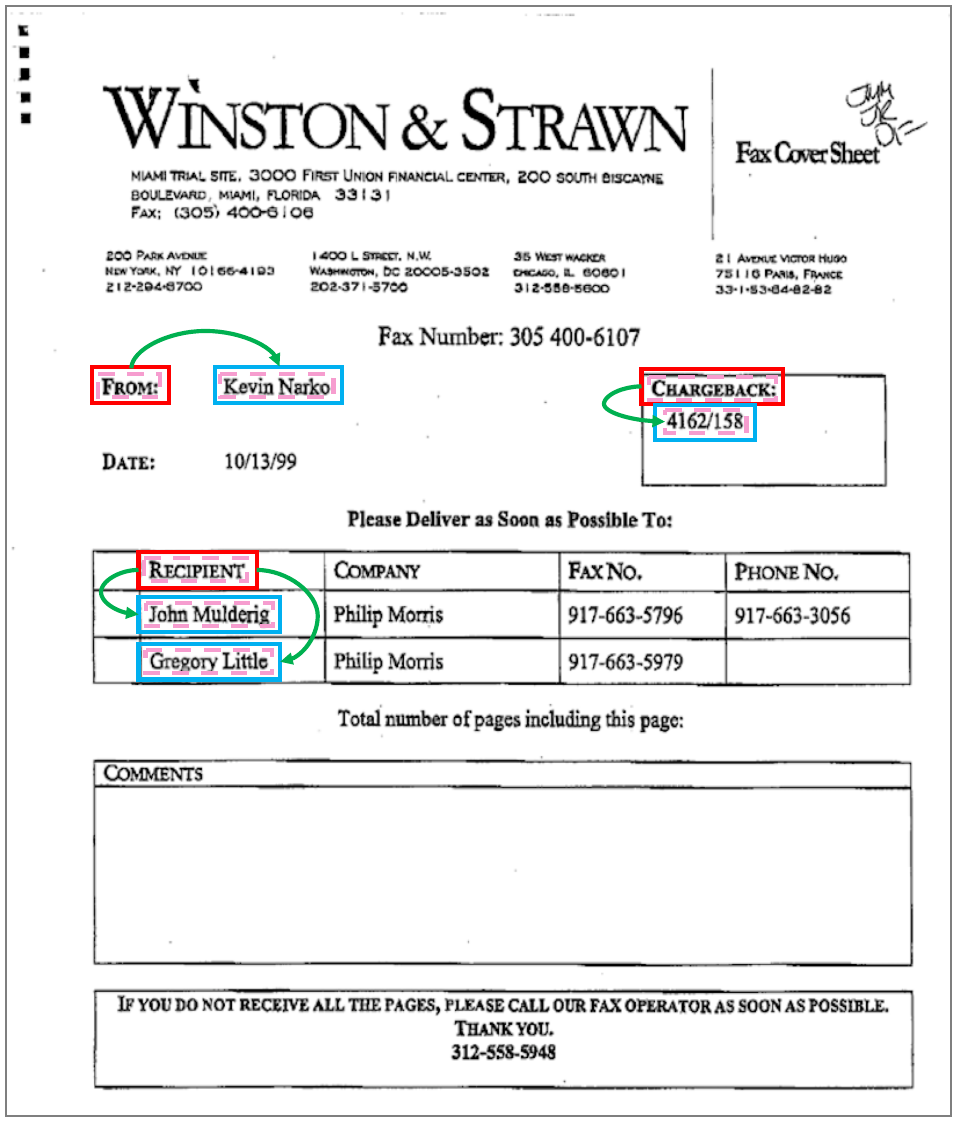}\label{fig:entity_linking}}\\
	\caption{Examples of VRDs and their key extraction information. The dotted boxes are the text regions and the solid ones are the semantic entity regions. (a) The entity extraction in token-level characters. (b) The entity extraction in segment-level text lines. (c) The relationship extraction with key-value pairs at segment-level.}
\label{fig:showcase}
\end{figure*}

Typical structure extraction methods rely on preliminary Optical Character Recognition (OCR) engines~\cite{zhang2019lomo,su2014accurate,wang2021pgnet,yu2020towards,wang2019sast,liao2020mask} to understand the semantics of documents. As shown in Figure~\ref{fig:showcase}, the contents in a document can be located as several text segments (pink dotted boxes) by text detectors. The entity fields are presented in three forms: partial characters, an individual segment, and multiple segment lines. Traditional methods for entity labeling often formulated the task as a sequential labeling problem. In this setup, the text segments are serialized as a linear sequence with a pre-defined order. Then a Named Entity Recognition (NER)~\cite{lample2016neural,MaH16} model is utilized to label each token such as word or character with an IOB (Inside, Outside, Beginning) tag. However, the capability is limited as the manner that is performed at token-level. As shown examples of Figure~\ref{fig:segment_labeling} and~\ref{fig:entity_linking}, VRDs are usually organized in a number of text segments. The segment-level textual content presents richer geometric and semantic information, which is vital for structured text understanding. Several methods~\cite{rener2020,wang2020docstruct,hwang2020spatial,guo2019eaten} focus on a segment-level representation. On the contrary, they cannot cope with the entity composed of characters as shown in Figure~\ref{fig:token_labeling}. Therefore, a comprehensive technique of structure extraction at both segment-level and token-level is worth considering.

Nowadays, accurate understanding of the structured text from VRDs remains a challenge. The key to success is the full use of multiple modal features from document images. Early solutions solve the entity tasks by only operating on plain texts, resulting in a semantic ambiguity. Noticing the rich visual information contained in VRDs, several methods~\cite{sarkhel2019visual,denk2019bertgrid,katti-etal-2018-chargrid,palm2019attend} exploit 2D layout information to provides complementation for textual content. Besides, for further improvement, mainstream researches~\cite{liu2019GCMIE,sage2020end,yu2020pick,ChengQSH020, majumder-etal-2020-representation,wang2021Ephoie,zhang2020trie} usually employ a shallow fusion of text, image, and layout to capture contextual dependencies. Recently, several pre-training models~\cite{xu2020layoutlm,pramanik2020towards,xu2020layoutlmv2} have been proposed for joint learning the deep fusion of cross-modality on large-scale data and outperform counterparts on document understanding. Although these pre-training models consider all modalities of documents, they focus on the contribution related to the text side with less elaborate visual features.

\begin{figure*}[t]
\begin{center}
\includegraphics[width=1.0\linewidth]{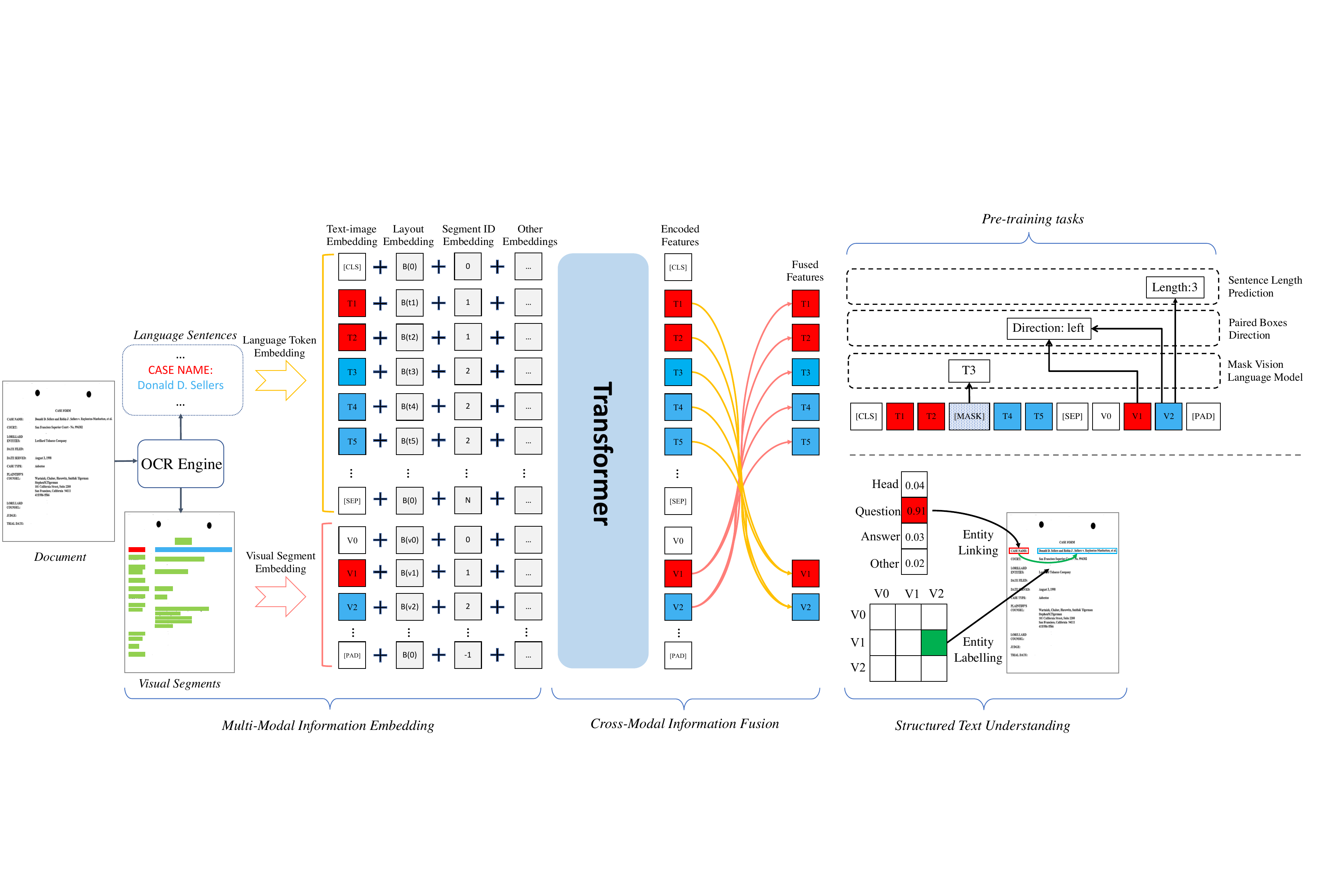}
\end{center}
\caption{An overall illustration of the model framework and the inform extraction tasks for StrucTexT.}
\label{fig:framework}
\end{figure*}

To address the above limitations, in this paper, we propose a uniform framework named \textbf{StrucTexT} that incorporates the features from different levels and modalities to effectively improves the understanding of various document structures. Inspired by recent developments in vision-language transformers~\cite{Lu2019ViLBERTPT,Su2020VLBERTPO}, we introduce a transformer encoder to learn cross-modal knowledge from both images of segments and tokens of words. In addition, we construct an extra segment ID embedding to associate visual and textual features at different granularity. Meanwhile, we attach a 2D position embedding to involve the layout clue. After that, a Hadamard product works on the encoded features between different levels and modalities for advanced feature fusion. Hence, StrucTexT can support segment-level and token-level tasks of structured text understanding in a single framework. Figure~\ref{fig:framework} shows the architecture of our proposed method.

To promote the representation capacity of multi-modality, we further introduce three self-supervised tasks for pre-training learning of text, image, and layout. Specifically, following the work of LayoutLM~\cite{xu2020layoutlm}, the Masked Visual Language Modeling (MVLM) task is utilized to extract contextual information. In addition, we present two tasks named Sentence Length Prediction (SLP) and Paired Boxes Direction (PBD). SLP task predicts the segment length for enhancing the internal semantics of an entity candidate. PBD task is training to identify the relative direction within a sampled segment pair, which helps our framework discover the geometric structure topology. The three self-supervised tasks make full use of both textual and visual features of the documents. An unsupervised pre-training strategy with above all tasks is applied at first to get an enhanced feature encoder.

Major contributions of this paper are summarized as follows:
\begin{enumerate}
\item In this paper, we present a novel framework named StrucTexT to tackle the tasks of structured text understanding with a unified solution. It efficiently extracts semantic features from different levels and modalities to handle the entity labeling and entity linking tasks.
\item  We introduce improved multi-task pre-training strategies to extract the multi-modal information from VRDs by self-supervised learning. In addition to the MVLM task that benefits the textual context, we proposed two new pre-training tasks of SLP and PBD to take advantage of image and layout features. We adopt the three tasks during the pre-training stage for a richer representation.
\item Extensive experiments on real-world benchmarks show the superior performance of StrucTexT compared with the state-of-the-art methods. Additional ablation studies demonstrate the effectiveness of our pre-training strategies.
\end{enumerate}

\section{Related Work}

\textbf{Structured Text Understanding} The task of structured text understanding is to retrieve structured data from VRDs automatically. It requires the model to extract the semantic structure of textual content robustly and effectively, assigning the major purpose into two parts~\cite{jaume2019funsd}: entity labeling and entity linking. Generally speaking, the entity labeling task is to find named entities. The entity linking task is to extract the semantic relationships as key-value pairs between entities. Most existing methods~\cite{hwang2019post,denk2019bertgrid,devlin2018bert, xu2020layoutlm, xu2020layoutlmv2, yu2020pick, zhang2020trie, wang2021Ephoie} design a NER framework to perform entity labeling as a sequence labeling task at token-level. However, traditional NER models organize text in one dimension depending on the reading order and are unsuitable for VRDs with complex layouts. Recent studies~\cite{qian-etal-2019-graphie,wei2020robust,yu2020pick,zhang2020trie,wang2021Ephoie,xu2020layoutlm,xu2020layoutlmv2} have realized the significance of segment-level features and incorporate a segment embedding to attach extra higher semantics. Although those methods, such as PICK~\cite{yu2020pick} and TRIE~\cite{zhang2020trie}, construct contextual features involving the segment clues, they revert to token-level labeling with NER-based schemes. Several works \cite{ChengQSH020,rener2020,wang2020docstruct,hwang2020spatial} design their methods at segment-level to solve the tasks of entity labeling and entity linking. \citet{ChengQSH020} utilizes an attention-based network to explore one-shot learning for the text field labeling. DocStruct~\cite{wang2020docstruct} predicts the key-value relations between the extracted text segments to establish a hierarchical document structure. With the graph-based paradigm, \citet{rener2020} and \citet{hwang2020spatial} tackle the entity labeling and entity linking tasks simultaneously. However, they don’t consider the situation where a text segment includes more than one category, which is difficult to identify the entity in token granularity.

In summary, the methods mentioned above can only handle one granularity representation. To this end, we propose a unified framework to support both token-level and segment-level structured extraction for VRDs. Our model is flexible to any granularity-modeling tasks for structured text understanding.

\noindent
\textbf{Multi-Modal Feature Representations} One of the most important modules of structured information extraction is to understand multi-modal semantic features. 
Previous works~\cite{lample2016neural,dai2019transformer,dengel2002smartfix,palm2017cloudscan,devlin2018bert,sage2019recurrent,hwang2019post} usually adopt language models to extract entities from the plain text. These NLP-based approaches typically operate on text sequences and do not incorporate visual and layout information.
Later studies~\cite{sarkhel2019visual,denk2019bertgrid,katti-etal-2018-chargrid} firstly tend to explore layout information to aid entity extraction from VRDs. 
Post-OCR~\cite{hwang2019post} reconstructs the text sequences based on their bounding boxes. 
VS2~\cite{sarkhel2019visual} leverages the heterogeneous layout to perform the extraction in visual logical blocks. 
A range of other methods~\cite{denk2019bertgrid,katti-etal-2018-chargrid,zhao2019cutie} represent a document as a 2D grid with text tokens to obtain the contextual embedding. After that, some researchers realize the necessity of multi-modal fusion and develop performance by integrating visual and layout information. GraphIE~\cite{qian-etal-2019-graphie}, PICK~\cite{yu2020pick} and \citet{liu2019GCMIE} design a graph-based decoder to improve the semantics of context information. \citet{hwang2020spatial} and \citet{wang2020docstruct} leverage the relative coordinates and explore the link of each key-value pair. These methods only use simple early fusion strategies, such as addition or concatenation, without considering the semantic gap of different modalities. Recently, pre-training models~\cite{devlin2018bert,sun2019ernie20,Lu2019ViLBERTPT,Su2020VLBERTPO} show a strong feature representation using large-scale unlabeled training samples. Inspired by this, several works~\cite{xu2020layoutlm,xu2020layoutlmv2,pramanik2020towards} combine pre-training techniques to improve multi-modal features. \citet{pramanik2020towards} introduces a multi-task learning-based framework to yield a generic document representation. LayoutLMv2~\cite{xu2020layoutlmv2} uses 11 million scanned documents to obtain a pre-trained model, which shows the state-of-the-art performance in several downstream tasks of document understanding. However, these pre-training strategies mainly focus on the expressiveness of language but underuse the structured information from images. Hence, we propose a self-supervised pre-training strategy to better explore the potentials information from text, image, and layout. Compared with LayoutLMv2, the new strategy supports more useful features with less training data.

\section{APPROACH}

Figure~\ref{fig:framework} shows the overall illustration of StrucTexT. Given an input image with preconditioned OCR results, such as bounding boxes and content of text segments. We leverage various information from text, image, and layout aspects by a feature embedding stage. And then, the multi-modal embeddings are fed into the pre-trained transformer network to obtain rich semantic features. The transformer network has accomplished the cross-modality fusion by establishing interactions between the different modality inputs. At last, the Structured Text Understanding module receives the encoded features and carries out entity recognition for entity labeling and relation extraction for entity linking.

%-------------------------------------------------------------------------
\subsection{Multi-Modal Feature Embedding}
\label{section_3.1}
Given a document image $I$ with $n$ text segments, we perform open source OCR algorithms~\cite{zhou2017east,shi2016end} to obtain the $i$-th segment region with the top-left and bottom-right bounding box $b_i=(x_0, y_0, x_1, y_1)$ and its corresponding text sentence $t_i=\{c^i_1, c^i_2, \cdots, c^i_{l_i}\}$, where $c$ is a word or character and $l_i$ is the length of $t_i$.

\vspace{0.15cm}
\noindent
\textbf{Layout Embedding} For every segment or word, we use the encoded bounding boxes as their layout information
\begin{equation}
\begin{aligned}
 \text{L} = \text{Emb}_l(x_0, y_0, x_1, y_1, w, h)
\end{aligned}
  \label{con:layout_emb}
\end{equation}
where $\text{Emb}_l$ is a layout embedding layer and $w, h$ is the shape of bounding box $b$. It is worth mentioning that we estimate the bounding box of a word by its belonging text segment in consideration of some OCR results without word-level information.

\vspace{0.15cm}
\noindent
\textbf{Language Token Embedding} Following the common practice~\cite{devlin2018bert}, we utilize the WordPiece~\cite{wu2016wordPiece} to tokenize text sentences. After that, all of text sentences are gathered as a sequence $\text{S}$ by sorting the text segments from the top-left to bottom-right. Intuitively, a pair of special tags [CLS] and [SEP] are added at the beginning and end of the sequence, as $t_0=\{\text{[CLS]\}}, t_{n+1}=\{\text{[SEP]\}}$. Thus, we can define the language sequence $\text{S}$ as follows
\begin{equation}
\begin{aligned}
 \text{S} & = \left\{t_0, t_1, \cdots, t_n, t_{n+1} \right\} \\
 & = \left\{\text{[CLS]}, c^1_1, \cdots, c^1_{l_1}, \cdots, c^n_1, \cdots, c^n_{l_n}, \text{[SEP]} \right\}
\end{aligned}
\end{equation}

Then, we sum the embedded feature of $S$ and layout embedding $L$ to obtain the language embedding $\text{T}$
\begin{equation}
\text{T} = \text{Emb}_t(\text{S}) + \text{L}
\end{equation}
where $\text{Emb}_t$ is a text embedding layer.

\vspace{0.15cm}
\noindent
\textbf{Visual Segment Embedding} In the model architecture, we use ResNet50~\cite{xie2017resnet} with FPN~\cite{lin2017fpn} as the image feature extractor to generate feature maps of $I$. Then, the image feature of each text segment is extracted from the CNN maps by RoIAlign~\cite{he2017mask} according to $b$. The visual segment embedding $\text{V}$ is computed as
\begin{equation}
 \text{V}=\text{Emb}_v(\text{ROIAlign}(\text{CNN}(I), b)) + \text{L}
\end{equation}
where $Emb_v$ is the visual embedding layer. Furthermore, the entire feature maps of image $I$ is embedded as $\text{V}_0$ to introduce the global information into image features.

\vspace{0.15cm}
\noindent
\textbf{Segment ID Embedding} Compared with the vision-language tasks based on wild pictures, understanding the structured document requires higher semantics to identify the ambiguous entities. Thus, we propose a segment ID embedding $S^{id}$ to allocate a unique number to each text segment with its image and text features, which makes an explicit alignment of cross-modality clues.

\vspace{0.15cm}
\noindent
\textbf{Other Embeddings} In addition, we add two other embeddings~\cite{Lu2019ViLBERTPT,Su2020VLBERTPO} into the input. The position embedding $P^{id}$ encodes the indexes from 1 to maximum sequence length, and the segment embedding $M^{id}$ denotes the modality for each feature. All above embeddings have the same dimensions. In the end, the input of our model is represented as the combination of the embeddings.

\begin{equation}
\begin{aligned}
      \text{Input} = \text{Concat}(T, V) + S^{id} + P^{id} + M^{id}
\end{aligned}
\end{equation}

Moreover, we append several [PAD] tags to fill the short input sequence to a fixed length. An empty bounding box with zeros is assigned to the special [CLS], [SEP], and [PAD] tags.

\begin{figure*}[t]
\begin{center}
   \includegraphics[width=0.85\linewidth]{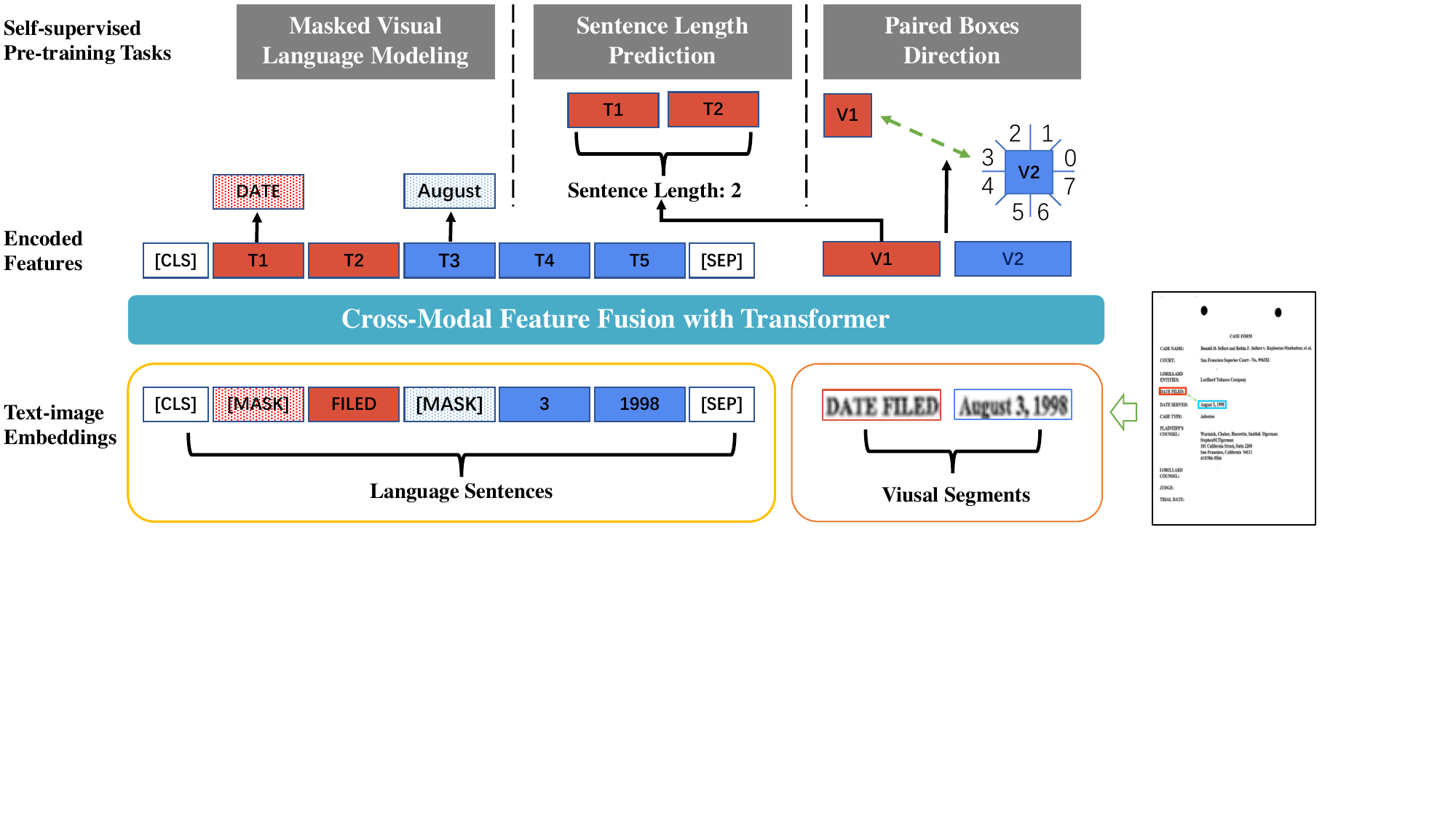}
\end{center}
  \caption{The illustration of cross-modal information fusion. Three self-supervised tasks MVLM, SLP, and PBD introduced in Section~\ref{section_3.1} are employed simultaneously on the visual and language embeddings in the pre-training stage.}
\label{fig:feature_fusion}
\end{figure*}

%-------------------------------------------------------------------------
\subsection{Multi-Modal Feature Enhance Module}

StrucText collects multi-modal information from visual segments, text sentences, and position layouts to produce an embedding sequence. We support an image-text alignment between different granularities by leveraging the segment IDs mentioned above. At this stage, we perform a transformer network to encode the embedding sequence to establish deep fusion between modalities and granularities. Crucially, three self-supervised tasks encode the input features during the pre-training stage to learn task-agnostic joint representations. The details are introduced as follows, where patterns of all self-supervised tasks are as shown in Figure~\ref{fig:feature_fusion}.

\vspace{0.15cm}
\noindent
\textit{Task 1: Masked Visual Language Modeling} 

\noindent
The MVLM task promotes capturing a contextual representation on the language side. Following the pattern of \textit{masked multi-modal modeling} in ViLBERT~\cite{Lu2019ViLBERTPT}, we select 15\% tokens from the language sequences, mask 80\% among them with a [MASK] token, replace 10\% of that with random tokens, and keep 10\% tokens unchanged. Then, the model is required to reconstruct the corresponding tokens. Rather than following the image region mask in ViLBERT, we retain all other information and encourage the model to hunt for the cross-modality clues at all possible.

\vspace{0.15cm}
\noindent
\textit{Task 2: Segment Length Prediction}

\noindent
Besides the MLVM, we introduce a new self-supervised task called Sequence Length Prediction (SLP) to excavate fine-grained semantic information on the image side. The SLP task asks the model to recognize the length of the text segment from each visual feature. In this way, we force the encoder to learn from the image feature, more importantly, the language sequence knowledge via the same segment ID. We argue that this information flow could accelerate the deep cross-modal fusion among textual, visual, and layout information. 

Moreover, to avoid the disturbance of sub-words produced by WordPiece~\cite{wu2016wordPiece}, we only count each first sub-word for keeping the same length between language sequences and image segments. Therefore, we build an extra alignment between two granularities, which is simple but effective.

\vspace{0.15cm}
\noindent
\textit{Task 3: Paired Box Direction}

\noindent
Furthermore, our third self-supervised task, Paired Box Direction (PBD), is designed to exploit global layout information. The PBD task aims at learning a comprehensive geometric topology for document structures by predicting the pairwise spatial relationships of text segments. First of all, we divide the field of 360 degrees into eight identical buckets. Secondly, we compute the angle $\theta_{ij}$ between the text segment $i$ and $j$ and label it with one of the buckets. Next, we carry out the subtraction between two visual features on the image side and take the result $\triangle \hat{V}_{ij}$ as the input of the PBD
\begin{equation}
     \triangle \hat{V}_{ij} = \hat{V}_{i} - \hat{V}_{j}
\end{equation}
where we use the $\hat{}$ symbol to denote the features after transformer encoding. $\hat{V}_{i}$ and $\hat{V}_{j}$ express the visual features for $i$-th segment and $j$-th segment.

Finally, we define PBD as a classification task to estimate the relative positional direction with $\triangle \hat{V}_{ij}$. 

%-------------------------------------------------------------------------
\subsection{Structural Text Understanding}

\vspace{0.15cm}
\noindent
\textbf{Cross-granularity Labeling Module} The cross-granularity labeling module supports both token-level entity labeling and segment-level entity labeling tasks. In this module, tokens with the same segment ID on the language side are aggregated into a segment-level textual feature through the arithmetic average
\begin{equation}
     \hat{T}_i = \text{mean}(\hat{t}_i) = (\hat{c}_{1} + \hat{c}_{2} + \cdots \hat{c}_{l_i}) / l_i
\end{equation}
where $\hat{t}_i$ means the features of $i$-th text sentence, $\hat{c}$ is the feature of token, $l_i$ is the sentence length. After that, a bilinear pooling layer is utilized to compute a Hadamard product to fuse the textual segment feature $T_i$ and the visual segment feature $V_i$.

\begin{equation}
    X_i = V_i * T_i
\end{equation}

Finally, we apply a fully connected layer on the cross-modal features $X_i$ to predict an entity label for segment $i$ with the Cross-Entropy loss.

\vspace{0.15cm}
\noindent
\textbf{Segment Relationship Extraction Module} The segment relationship extraction module is proposed for entity linking. Documents usually represent their structure as a set of hierarchical relations, such as key-value pair or table parsing. Inspired by Docstruct~\cite{wang2020docstruct}, we use an asymmetric parameter matrix $M$ to extract the relationship from segments $i$ to $j$ in probability form
 
\begin{equation}
    P_{i\rightarrow j}= \sigma(X_\text{j}MX_{i}^{T})
\end{equation}
where $P_{i\rightarrow j}$ is the probability of whether $i$ links to $j$. $M$ is a parameter matrix and $\sigma$ is the sigmoid function.

We notice that most of the segment pairs in a document are not related. To alleviate the data sparsity and balance the number of related and unrelated pairs, we learn from the Negative Sampling method~\cite{mikolov2013distributed} and build a sampling set with non-fixed size. Our sampling set consists of the same number of positive and negative samples.

However, we also find the training process is unstable only using the above sampling strategy. To utmost handle the imbalanced distribution of entity linking, we combine the Margin Ranking Loss and Binary Cross-Entropy to supervise the training simultaneously. Thus, the linking loss can be formulated as
\begin{equation}
    \text{Loss} = \text{Loss}_\text{BCE} + \text{Loss}_\text{Rank}
\end{equation}
where the $\text{Loss}_\text{Rank}$ is computed as following

\begin{equation}
    \text{Loss}_\text{Rank}(P_i, P_j, y) =  max(0, -y * (P_i - P_j) + \text{Margin}),
\end{equation}

Note that $y$ equals 1 if $(P_i, P_j)$ is the positive-negative samples pair or equals 0 for the negative-positive samples pair.

\section{EXPERIMENTS}
\subsection{Datasets}

In this section, we firstly introduce several datasets that are used for pre-training and evaluating our StrucTexT. The extensive experiments are conduct on three benchmark databases: FUNSD~\cite{jaume2019funsd}, SROIE~\cite{huang2019sroie}, EPHOIE~\cite{wang2021Ephoie}. Moreover, we perform ablation studies to analyze the effects of each proposed component.

\vspace{0.15cm}
\noindent
\textbf{DOCBANK}~\cite{li2020docbank}
contains 500K document pages (400K for training, 50K for validation and 50K for testing) for document layout analysis. We pre-train StrucTexT on the dataset.

\vspace{0.15cm}
\noindent
\textbf{RVL-CDIP}~\cite{harley2015cdip}
consists of 400,000 grayscale images in 16 classes, with 25,000 images per class. There are 320,000 training images, 40,000 validation images and 40,000 test images. We adopt RVL-CDIP for pre-training our model.

\vspace{0.15cm}
\noindent
\textbf{FUNSD}~\cite{jaume2019funsd}
consists of 199 real, fully annotated, scanned form images.
The dataset is split into 149 training samples and 50 testing samples. Three sub-tasks (word grouping, semantic entity labeling, and entity linking) are proposed to identify the semantic entity (i.e., questions, answers, headers, and other) and entity linking present in the form. We use the official OCR annotation and focus on the latter two tasks in this paper.

\vspace{0.15cm}
\noindent
\textbf{SROIE}~\cite{huang2019sroie}
is composed of 626 receipts for training and 347 receipts for testing. Every receipt contains four predefined values: company, date, address, and total. The segment-level text bounding box and the corresponding transcript are provided according to the annotations. We use the official OCR annotations and evaluate our model for receipt information extraction.

\vspace{0.15cm}
\noindent
\textbf{EPHOIE}~\cite{wang2021Ephoie}
is collected from actual Chinese examination papers with the diversity of text types and layout distribution. The 1,494 samples are divided into a training set with 1,183 images and a testing set with 311 images, respectively. Every character in the document is annotated with a label from ten predefined categories. The token-level entity labeling task is evaluated in this dataset.

\begin{table}
    \begin{tabular}{l | c | c | c | c}
    \hline
    Method         & Prec. & Recall & F1 & Params. \\ 
    \hline
    LayoutLM\_\text{BASE}~\cite{xu2020layoutlm} & 94.38 & 94.38 & 94.38 & 113M\\
    LayoutLM\_\text{LARGE}~\cite{xu2020layoutlm} & 95.24 & 95.24 & 95.24 & 343M\\
    PICK~\cite{yu2020pick}         & \textbf{96.79} & 95.46 & 96.12 & -\\
    VIES~\cite{wang2021Ephoie}     & - & - & 96.12 & -\\
    TRIE~\cite{zhang2020trie}      & - & - & 96.18 & -\\
    LayoutLMv2\_\text{BASE}~\cite{xu2020layoutlmv2} & 96.25 & 96.25 & 96.25 & 200M\\
    MatchVIE~\cite{tang2021matchvie} & - & - & 96.57 & -\\
    LayoutLMV2\_\text{LARGE}~\cite{xu2020layoutlmv2} & 96.61 & 96.61 & 96.61 & 426M\\
    \hline
    Ours & 95.84 & \textbf{98.52} & \textbf{96.88} & 107M\\
    & & & ($\pm$0.15) \\
    \hline
    \end{tabular}
    \centering
	\vspace{0.1cm}\caption{Model Performance (entity labeling) comparison on the SROIE dataset.}
	\label{tab:sroie_entity_labeling_res}
\end{table}

\begin{table}
    \begin{tabular}{l | c | c | c | c}
    \hline
    Method         & Prec. & Recall & F1 & Params. \\
    \hline
    \citet{rener2020} & - & - & 64.0 & -\\
    SPADE~\cite{hwang2020spatial} & - & - & 70.5 & -\\
    LayoutLM\_\text{BASE}~\cite{xu2020layoutlm} & 75.97 & 81.55 & 78.66 & 113M\\
    LayoutLM\_\text{LARGE}~\cite{xu2020layoutlm} & 75.96 & 82.19 & 78.95 & 343M\\
    MatchVIE~\cite{tang2021matchvie} & - & - & 81.33 & - \\
    LayoutLMv2\_\text{BASE}~\cite{xu2020layoutlmv2} & 80.29 &85.39 & 82.76 & 200M\\
    LayoutLMv2\_\text{LARGE}~\cite{xu2020layoutlmv2} & 83.24 &\textbf{85.19} & \textbf{84.20} & 426M\\
    \hline
    Ours & \textbf{85.68} & 80.97 & 83.09 & 107M\\
    & & & ($\pm$0.09) \\
    \hline
    \end{tabular}
    \centering
	\vspace{0.1cm}\caption{Model Performance (entity labeling) comparison on the FUNSD dataset, We ignore entities belonging to the \textit{other} category and use the mean performance of three classes (header, question, and answer) as our final results.}
	\label{tab:funsd_entity_labeling_res}
\end{table}

\begin{table}
    \begin{tabular}{ l | p{0.6cm} | p{0.75cm} | p{0.68cm} | p{0.68cm} | p{0.68cm} | p{0.45cm}}
    \hline
    \multirow{2}{*}{Method} & \multicolumn{2}{c|}{Reconstruction} & \multicolumn{3}{c|}{Detection} & \multirow{2}{*}{F1} \\ \cline{2-6}
                  & mAP & mRank & Hit@1 & Hit@2 & Hit@5  \\ 
     \hline
     FUNSD~\cite{jaume2019funsd}        & - & - & - & - & - & 4.0 \\ 
     \citet{rener2020}       & - & - & - & - & - & 39.0 \\
     LayoutLM$^*$~\cite{xu2020layoutlm} & 47.61 & 7.11 & 32.43 & 45.56 & 66.41 & -   \\
     DocStruct~\cite{wang2020docstruct}    & 71.77 & \textbf{2.89} & 58.19 & 76.27 & 88.94 & -   \\ 
     SPADE~\cite{hwang2020spatial}    &  - & - & - & - & - & 41.7   \\
     \hline
     Ours & \textbf{78.36} & 3.38  & \textbf{67.67}  & \textbf{84.33} & \textbf{95.33} & \textbf{44.1} \\
    \hline
    \end{tabular}
    \centering
	\vspace{0.1cm}\caption{Model Performance (entity linking) comparison on the FUNSD dataset. (LayoutLM$^*$ is implemented by~\cite{wang2020docstruct})}
	\label{tab:funsd_entity_linking_res}
\end{table}

\begin{table*}[htp]
    \newcommand{\tabincell}[2]{\begin{tabular}{@{}#1@{}}#2\end{tabular}}
    \begin{tabular}{l | c | c | c | c | c | c | c | c | c | c | c}
    \hline
     \multirow{2}{*}{\textbf{Method}}  &\multicolumn{11}{c}{\textbf{Entities}}
    \\ \cline{2-12}
     & Subject & \tabincell{c}{Test \\ Time} & Name & School & \tabincell{c}{Exam \\ Number} & \tabincell{c}{Seat \\ Number} & Class & \tabincell{c}{Student \\ Number} & Grade & Score & Mean \\ 
     \hline
     TRIE~\cite{zhang2020trie} &98.79 &\textbf{100} &99.46 &99.64 &88.64 &85.92 &97.94 &84.32 &97.02 &80.39 &93.21 \\
     VIES~\cite{wang2021Ephoie} &99.39 &\textbf{100} &99.67 &99.28 &91.81 &88.73 &99.29 &89.47 &98.35 &86.27 &95.23 \\
     MatchVIE~\cite{tang2021matchvie} &\textbf{99.78} &\textbf{100} &\textbf{99.88} &98.57 &94.21 &93.48 &\textbf{99.54} &92.44 &98.35 &92.45 &96.87 \\
     \hline
     Ours &99.25 &\textbf{100} &99.47  &\textbf{99.83} &\textbf{97.98} &\textbf{95.43} &98.29 &\textbf{97.33} &\textbf{99.25} &\textbf{93.73} &\textbf{97.95} \\
    \hline
    \end{tabular}
    \centering
	\vspace{0.1cm}\caption{Model Performance (token-level entity labeling) comparison on the EPHOIE dataset.}
	\label{tab:ephoie_entity_labeling_res}
\end{table*}

\subsection{Implementation}

\noindent
Following the typical pre-training and fine-tuning strategies, we train the model end-to-end. Across all pre-training and downstream tasks, we rescale the images and pad them to the size of $512\times 512$. The input sequence is set as a maximum length of 512.

\subsubsection{Pre-training}

\noindent
We extract both token-level text features and segment-level visual features based on a unified joint model by the encoder. Due to time and computational resource restrictions, we choose the 12-layer transformer encoder with 768 hidden size and 12 attention heads. We initialize the transformer and the text embedding layer from the ERNIE$_\text{BASE}$~\cite{sun2019ernie20}. The weights of the ResNet50 network is initialized using the ResNet\_vd~\cite{he2019bag} pre-trained on the ImageNet~\cite{deng2009imagenet}. The rest of the parameters are randomly initialized.

To obtain the pre-training OCR results, we apply the PaddleOCR~\footnote{https://github.com/PaddlePaddle/PaddleOCR} to extract the text segment in both DOCBANK and RVL-CDIP datasets. All three self-supervised tasks are trained for classification with the Cross-Entropy loss. The Adamax optimizer is used with an initial $5 \times 10^{-5}$ learning rate for a warm-up. And then, we keep $1 \times 10^{-4}$ for 2$\sim$5 epochs and set a linear decay schedule in the rest of epochs. We pre-train our architecture in DOCBANK~\cite{li2020docbank} and RVL-CDIP~\cite{harley2015cdip} dataset for 10 epochs with a batch size of 64 on 4 NVIDIA Tesla V100 32GB GPUs.

\subsubsection{Fine-tuning}

We fine-tune our StrucText on three information extraction tasks: entity labeling and entity linking at segment-level and entity labeling at token-level. For the segment-based entity labeling task, we aggregate token features of the text sentence via the arithmetic average and get the segment-level features by multiplying visual features and textual features. At last, a softmax layer is followed by the features for segment-level category prediction. The entity-level F1-score is used as the evaluation metric.

The entity linking task takes two segment features as input to obtain a pairwise relationship matrix. Then we pass the non-diagonal elements in the relationship matrix through a sigmoid layer to predict the binary classification of each relationship.

For the token-based entity labeling task, the output visual feature is expanded as a vector with the same length of its text sentence. Next, the extended visual features are element-wise multiplied with the corresponding textual features to obtain token-level features to predict the category of each token through a softmax layer.

We fine-tune our pre-trained model at all downstream tasks for 50 epochs with a batch size of $4$ and a learning rate from $1 \times 10^{-4}$ to $1 \times 10^{-5}$. We use the precision, recall, and F1-score as evaluation metrics for entity labeling. Following DocStruct~\cite{wang2020docstruct} and SPADE~\cite{hwang2020spatial}, the performance of entity linking is estimated with Hit@1, Hit@2, Hit@5, mAp, mRank, and F1-score.

\subsection{Comparison with the State-of-the-Arts}

\noindent
We evaluate our proposed StrucTexT in three publish benchmarks for both the entity labeling and entity linking tasks.

\noindent
\textbf{Segment-level Entity Labeling}
The comparison results are shown in Table~\ref{tab:sroie_entity_labeling_res}. We can observe that StrucText exhibits a superior performance over baseline methods~\cite{tang2021matchvie,wang2021Ephoie,xu2020layoutlm,xu2020layoutlmv2,yu2020pick,zhang2020trie} on SROIE. Specifically, our method obtains a precision of 95.84\% and a Recall of 98.52\% in SROIE, which surpass that of LayoutLMv2\_\text{LARGE}~\cite{xu2020layoutlmv2} by 0.27\% in F1-score.

As shown in Table~\ref{tab:funsd_entity_labeling_res}, our method achieves competitive F1-score of 83.09\% in FUNSD. Although LayoutLMv2\_\text{LARGE} beats our F1-score by \~1\%, it is worth noting that LayoutLMv2\_\text{LARGE} using a larger transformer consisting of 24 layers and 16 heads that contains 426M parameters. Further, our model using only 90K documents for pre-training compared to LayoutLMv2\_\text{LARGE} which uses 11M documents. On the contrary, our model shows a better performance than LayoutLMv2\_\text{BASE} under the same architecture settings. It fully proves the superiority of our proposed framework. Moreover, to verify the performance gain is statistically stable and significant, we repeat our experiments five times to eliminate the random fluctuations and attach the standard deviation below the F1-score. 

\vspace{0.1cm}
\noindent
\textbf{Segment-level Entity Linking}
As shown in Table~\ref{tab:funsd_entity_linking_res}, we compare our method with several state-of-the-arts on FUNSD for entity linking. The baseline method~\cite{jaume2019funsd} gets the remarkably worst result with a simple binary classification for pairwise entities. The SPADE~\cite{hwang2020spatial} shows a tremendous gain by leading a Graph into their model. Compared with the SPADE, our method has a 2.4\% improvement and achieves 44.1\% F1-score. Besides, we evaluate the performance in the mAp, mRank, and Hit metrics mentioned in DocStruct~\cite{wang2020docstruct}. Our method attains 78.36\% mAP, 79.19\% Hit@1, 84.33\% Hit@2, and 95.33\% Hit@5 which outperforms DocStruct and obtains a competitive performance at the 3.38 mRank score.

\vspace{0.1cm}
\noindent
\textbf{Token-level Entity Labeling}
We further perform StrucText on EPHOIE. It is noticed that the entities annotated in this dataset are character-based. Therefore, we apply our StrucText to calculate the entities with the token-level prediction. Table~\ref{tab:ephoie_entity_labeling_res} illustrates the overall performance of the EPHOIE dataset. Our StrucText contributes to a top-tier performance with 97.95\%.

\subsection{Ablation Study}

We study the impact of individual components in StrucText and conduct ablation studies on the FUNSD and SROIE datasets.

\begin{table}[htp]
    \newcommand{\tabincell}[2]{\begin{tabular}{@{}#1@{}}#2\end{tabular}}
    \begin{tabular}{c | c | c | c | c}
    \hline
    Dataset & \tabincell{c}{Pre-training \\ Tasks} & Prec. & Recall & F1     \\ 
    \hline
    \multirow{2}{*}{FUNSD} & MVLM & 76.41 & 79.36 & 77.71 \\
        & MVLM+PBD & 81.22 & 79.46 & 80.29 \\
        & MVLM+SLP & \textbf{87.45} & 78.69 & 82.12 \\
        & MVLM+PBD+SLP & 85.68 & \textbf{80.97} & \textbf{83.09} \\
    \hline
    \multirow{2}{*}{SROIE} & MVLM & 95.25 & 97.89 & 96.54 \\
        & MVLM+PBD & 95.32 & 98.25 & 96.75 \\
        & MVLM+SLP & 95.30 & 98.16 & 96.70 \\
        & MVLM+PBD+SLP & \textbf{95.84} & \textbf{98.52} & \textbf{96.88} \\
    \hline
    \end{tabular}
    \centering
	\vspace{0.1cm}\caption{Ablation Studies with entity labeling on the FUNSD and SROIE datasets.}
	\label{tab:abl_labeling}
\end{table}
\begin{table}[htp]
    \begin{tabular}{c | c | c | c | c}
    \hline
    Dataset & Modality & Prec. & Recall & F1 \\ 
    \hline
    \multirow{2}{*}{FUNSD}  & Visual &76.93  & 77.51 & 77.22 \\
        & Language & 81.73 & 79.38 & 80.49 \\
        & Visual + Language & \textbf{85.68} & \textbf{80.97} & \textbf{83.09} \\
    \hline
    \multirow{2}{*}{SROIE} & Visual &90.14 & 92.11 & 91.11 \\
        & Language &94.54 & 97.91 & 96.18 \\
         & Viusal + Language &\textbf{95.84} & \textbf{98.52} & \textbf{96.88} \\
    \hline
    \end{tabular}
    \centering
	\vspace{0.1cm}\caption{Ablation studies with visual-only and language-only entity labeling on the FUNSD and SROIE datasets.}
	\label{tab:abl_vis_lan}
\end{table}

\begin{figure}[t]
	\centering
	\small
	\subfloat[Cases from SROIE labeling]{\includegraphics[width=0.98\linewidth, trim=0 12.8cm 17.6cm 0, clip]{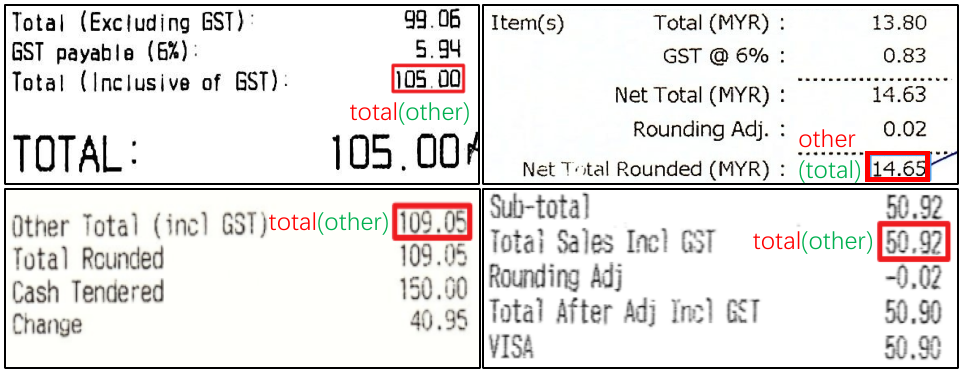}\label{fig:labeling_cases_of_sroie}} \\ 
	\subfloat[Cases from EPHOIE labeling]{\includegraphics[width=0.49\linewidth, trim=0 7.7cm 25.7cm 6.3cm, clip]{images/Framework_v2_structext_7.pdf}\label{fig:labeling_cases_of_ephoie}}
	\subfloat[Cases from FUNSD labeling]{\includegraphics[width=0.49\linewidth, trim=8.2cm 7.7cm 17.5cm 6.3cm, clip]{images/Framework_v2_structext_7.pdf}\label{fig:labeling_cases_of_funsd}}\\
	\subfloat[Cases from FUNSD linking]{\includegraphics[width=0.7\linewidth, trim=0.1cm 0.7cm 22.0cm 11.4cm, clip]{images/Framework_v2_structext_7.pdf}\label{fig:linking_cases_of_funsd}}\\
	\caption{Visualization of badcases of StrucText. For the entity labeling task in (a), (b), and (c), the wrong cases are shown as the red boxes (the correct results are hidden) and their nearby text represents the predications and ground-truths in red and green color, respectively. For the entity linking task in (d), the green/purple lines indicate the correct/error predicted linkings.}
\label{fig:case_study}
\end{figure}

\begin{table}[htp]
    \begin{tabular}{c | c | c | c | c}
    \hline
    Dataset & Granularity & Prec. & Recall & F1 \\ 
    \hline
    \multirow{2}{*}{FUNSD} & Token & 81.20  & \textbf{82.10} & 81.59 \\
        & Segment & \textbf{85.68} & 80.97 & \textbf{83.09} \\
    \hline
    \multirow{2}{*}{SROIE} & Token & 92.77 & \textbf{98.81} & 95.62 \\
        & Segment & \textbf{95.84} & 98.52 & \textbf{96.88} \\
    \hline
    \end{tabular}
    \centering
	\vspace{0.1cm}\caption{Ablation studies with the comparison of token-level and segment-level entity labeling on the FUNSD and SROIE datasets.}
	\label{tab:abl_token_seg}
\end{table}

\noindent
\textbf{Self-supervised Tasks in Pre-training}
In this study, we evaluate the impact of different pre-training tasks. As shown in Table ~\ref{tab:abl_labeling}, we can observe that the PBD and SLP tasks make better use of visual information. Specifically, compared with the model only trained with the MVLM task, MVLM+PBD gains nearly 3\% improvement in FUNSD and 0.2\% improvement in SROIE. Meanwhile, the results turn out that the SLP task also improves the performance dramatically. Furthermore, the incorporation of the three tasks obtains the optimal performance compared with other combinations. It means that both the SLP and PBD tasks contribute to richer semantics and potential relationships between cross-modality.

\noindent
\textbf{Multi-Modal Features Profits}
As shown in Table ~\ref{tab:abl_vis_lan}, then we perform experiments in verifying the benefits of features in multiple modalities. The textual features perform better than visual ones, which we attribute more semantics to the textual content of documents. Moreover, combining visual and textual features can achieve higher performance with a notable gap, indicating complementarity between language and visual information. The results show that the multi-modal feature fusion in our model can get a richer semantic representation.

\noindent
\textbf{Granularity of Feature Fusion}
We also study the representations with different granularities towards the performance. In detail, we complete the experiments of entity labeling on SROIE and FUNSD in token-level supervision. As shown in Table ~\ref{tab:abl_token_seg}, overall, the segment-based results perform better than token-based ones, which proves our opinion that the effectiveness of text segment.

\subsection{Error Analysis}
Although our work has achieved outstanding performance, we also observe some badcases of the proposed method. This section presents an error analysis on the qualitative results in SROIE, FUNSD and EPHOIE, respectively. In Figure~\ref{fig:labeling_cases_of_sroie}, our model makes the mistakes of wrong answers to the total in SROIE. We attribute the errors to the similar semantics of textual contents and the close distance of locations. In addition, as shown in Figure~\ref{fig:labeling_cases_of_ephoie}, our model is confused by the similar style of digits, which demonstrates the relatively low performance in the numeral entities in EPHOIE, such as exam number, seat number, student number, and score in~\ref{tab:ephoie_entity_labeling_res}. However, these entities can certainly be distinguished by their keywords. To this end, a goal-directed information aggregation of key-value pair is well worth considering, and we would study it in future works. As shown in Figure~\ref{fig:labeling_cases_of_funsd}, the model is failed in recognizing the header and the question in FUNSD. We analyze the model is overfitting the layout position of training data. Then, according to Figure~\ref{fig:linking_cases_of_funsd}, some links are assigned incorrectly. We attribute the errors to ambiguous semantics of relationships.

\section{Conclusion}
In this paper, we further explore improving the understanding of document text structure by using a unified framework. Our framework shows superior performance on three real-world benchmark datasets after applying novel pre-training strategies for the multi-modal and multi-granularity feature fusion. Moreover, we evaluate the influence of different modalities and granularities on the ability of entity extraction, thus providing a new perspective to study the problem of structured text understanding.



\bibliographystyle{ACM-Reference-Format}

\balance
\bibliography{sample-base}


\end{document}